\documentclass[12pt,fleqn]{article}
\usepackage{amssymb,amsmath}
\usepackage{graphicx}
\usepackage{times}
\usepackage{color}
\usepackage{lipsum}
\usepackage{setspace}

\title{Learning Precise Spike Train to Spike Train Transformations in
	Multilayer Feedforward Neuronal Networks\footnote{To appear in Neural
    Computation 2016}}
\author{
 Arunava Banerjee\\
 Computer and Information Science and Engineering Department\\
 University of Florida\\
 Gainesville, FL 32611-6120\\
\texttt{arunava@cise.ufl.edu}}
\date{}

\begin{document}
\footnotesep20pt
\maketitle
\titlepage

\begin{abstract}

We derive a synaptic weight update rule for learning temporally precise spike
train to spike train transformations in multilayer feedforward networks of
spiking neurons. The framework, aimed at seamlessly generalizing error
backpropagation to the deterministic spiking neuron setting, is based strictly
on spike timing and avoids invoking concepts pertaining to spike rates or
probabilistic models of spiking. The derivation is founded on two innovations.
First, an error functional is proposed that compares the spike train emitted by
the output neuron of the network to the desired spike train by way of their
putative impact on a virtual postsynaptic neuron. This formulation sidesteps
the need for spike alignment and leads to closed form solutions for all
quantities of interest. Second, virtual assignment of weights to spikes rather
than synapses enables a perturbation analysis of individual spike times and
synaptic weights of the output as well as all intermediate neurons in the
network, which yields the gradients of the error functional with respect to the
said entities. Learning proceeds via a gradient descent mechanism that
leverages these quantities. Simulation experiments demonstrate the efficacy of
the proposed learning framework. The experiments also highlight asymmetries
between synapses on excitatory and inhibitory neurons.

\end{abstract}

\section{Introduction}

In many animal sensory pathways, information about external stimuli is encoded
in precise patterns of neuronal spikes \cite{1,2,3,4,5,6}. If the integrity of
this form of information is to be preserved by downstream neurons, they have to
respond to these precise patterns of input spikes with appropriate, precise
patterns of output spikes. How networks of neurons can learn such spike train
to spike train transformations has therefore been a question of significant
interest. When the transformation is posited to map mean spike rates to mean
spike rates, error backpropagation \cite{7,8,9} in multilayer feedforward
networks of rate coding model of neurons has long served as the cardinal
solution to this learning problem. Our overarching objective in this article is
to develop a counterpart for transformations that map precise patterns of input
spikes to precise patterns of output spikes in multilayer feedforward networks
of deterministic spiking neurons, in an online setting. In particular, we aim
to devise a learning rule that is strictly spike timing based, that is, one
that does not invoke concepts pertaining to spike rates or probabilistic models
of spiking, and that seamlessly generalizes to multiple layers.

In an online setting, the spike train to spike train transformation learning
problem can be described as follows. At one's disposal is a spiking neuron
network with adjustable synaptic weights. The external stimulus is assumed to
have been mapped---via a fixed mapping---to an input spike train. This input
spike train is to be transformed into a desired output spike train using the
spiking neuron network. The goal is to derive a synaptic weight update rule
that when applied to the neurons in the network, incrementally brings the
output spike train of the network into alignment with the desired spike train.
With biological plausibility in mind, we also stipulate that the rule not
appeal to computations that would be difficult to implement in neuronal
hardware. We do not address the issue of what the desired output spike train in
response to an input spike train is, and how it is generated. We assume that
such a spike train exists, and that the network learning the transformation has
access to it. Finally, we do not address the question of whether the network
has the intrinsic capacity to implement the input/output mapping; we undertake
to learn the mapping without regard to whether or not the network, for some
settings of its synaptic weights, can instantiate the input/output
transformation.\footnote{Our goal is to achieve convergence for those mappings
that can be learned. For transformations that, in principle, lie beyond the
capacity of the network to represent, the synaptic updates are, by
construction, designed not to converge.} There is, at the current time, little
understanding of what transformations feedforward networks of a given
depth/size and of a given spiking neuron model can implement, although some
initial progress has been made in \cite{ramaswamy2014}.

\section{Background}

The spike train to spike train transformation learning problem, as described
above, has been a question of active interest for some time. Variants of the
problem have been analyzed and significant progress has been achieved over the
years.

One of the early results was that of the {\em SpikeProp} supervised learning
rule \cite{bohte2002}. Here a feedforward network of spiking neurons was
trained to generate a desired pattern of spikes in the output neurons, in
response to an input spike pattern of bounded length. The caveat was that each
output neuron was constrained to spike exactly once in the prescribed time
window during which the network received the input. The network was trained
using gradient descent on an error function that measured the difference
between the actual and the desired firing time of each output neuron. Although
the rule was subsequently generalized in \cite{booij2005} to accommodate
multiple spikes emitted by the output neurons, the error function remained a
measure of the difference between the desired and the first emitted spike of
each output neuron.

A subsequent advancement was achieved in the {\em Tempotron} \cite{gutig2006}.
Here, the problem was posed in a supervised learning framework where a spiking
neuron was tasked to discriminate between two sets of bounded length input
spike trains, by generating an output spike in the first case and remaining
quiescent in the second. The tempotron learning rule implemented a gradient
descent on an error function that measured the amount by which the maximum
postsynaptic potential generated in the neuron, during the time the neuron
received the input spike train, deviated from its firing threshold. Operating
along similar lines and generalizing to multiple desired spike times, the FP
learning algorithm \cite{memmesheimer2014} set the error function to reflect
the earliest absence (presence) of an emitted spike within (outside) a finite
tolerance window of each desired spike.

Elsewhere, several authors have applied the Widrow-Hoff learning rule by first
converting spike trains into continuous quantities, although the rule's
implicit assumption of linearity of the neuron's response makes its application
to the spiking neuron highly problematic, as explored at length in
\cite{memmesheimer2014}. For example, the {\em ReSuMe} learning rule for a
single neuron was proposed in \cite{ponulak2010} based on a linear-Poisson
probabilistic model of the spiking neuron, with the instantaneous output firing
rate set as a linear combination of the synaptically weighted instantaneous
input firing rates. The output spike train was modeled as a sample draw from a
non-homogeneous Poisson process with intensity equal to the variable output
rate. The authors then replaced the rates with spike trains. Although the rule
was subsequently generalized to multilayer networks in \cite{sporea2013}, the
linearity of the neuron model is once again at odds with the proposed
generalization.\footnote{When the constituent units are linear, any multilayer
network can be reduced to a single layer network. This also emerges in the
model in \cite{sporea2013} where the synaptic weights of the intermediate layer
neurons act merely as multiplicative factors on the synaptic weights of the
output neuron.} Likewise, the {\em SPAN} learning rule proposed in
\cite{mohemmed2012} convolved the spike trains with kernels (essentially,
turning them into pseudo-rates) before applying the Widrow-Hoff update rule.

A bird's eye view brings into focus the common thread that runs through these
approaches. In all cases there are three quantities at play: the prevailing
error $E(\cdot)$, the output $O$ of the neuron, and the weight $W$ assigned to
a synapse. In each case, the authors have found a {\em scalar quantity}
$\tilde{O}$ that stands-in for the real output spike train $O$: the timing of
the only/first spike in \cite{bohte2002,booij2005}, the maximum postsynaptic
potential/timing of the first erroneous spike in the prescribed window in
\cite{gutig2006,memmesheimer2014}, and the current instantaneous firing
rate/pseudo-rate in \cite{ponulak2010,sporea2013,mohemmed2012}. This has
facilitated the computation of $\partial E/ \partial \tilde{O}$ and $\partial
\tilde{O}/ \partial W$, quantities that are essential to implementing a
gradient descent on $E$ with respect to $W$.

Viewed from this perspective, the immediate question becomes why not address
$O$ directly instead of its surrogate $\tilde{O}$? After all, $O$ is merely a
vector of output spike times. Two major hurdles emerge upon reflection.
Firstly, $O$, although a vector, can be potentially unbounded in length.
Secondly, letting $O$ be a vector requires that $E(\cdot)$ compare the vector
$O$ to the desired vector of spike times, and return a measure of disparity.
This can potentially involve aligning the output to the desired spike train
which not only makes differentiating $E(\cdot)$ difficult, but also strains
biological plausibility \cite{florian2012}.

We overcome these issues in stages. We first turn to the neuron model and
resolve the first problem. We then propose a closed form differentiable error
functional $E(\cdot)$ that circumvents the need to align spikes. Finally,
virtual assignment of weights to spikes rather than synapses allows us to
conduct a perturbation analysis of individual spike times and synaptic weights
of the output as well as all intermediate neurons in the network. We derive the
gradients of the error functional with respect to all output and intermediate
layer neuron spike times and synaptic weights, and learning proceeds via a
gradient descent mechanism that leverages these quantities. The perturbation
analysis is of independent interest, in that it can be paired with other
suitable differentiable error functionals to devise new learning rules. The
overall focus on individual spike times, both in the error functional as well
as in the perturbation analysis, has the added benefit that it sidesteps any
assumptions of linearity in the neuron model or rate in the spike trains,
thereby affording us a learning rule for multilayer networks that is
theoretically concordant with the nonlinear dynamics of the spiking neuron.

\section{Model of the Neuron}

Our approach applies to a general setup where the membrane potential of a
neuron can be expressed as a sum of multiple weighted $n$-ary functions of
spike times, for varying $n$ (modeling the interactive effects of spikes),
where gradients of the said functions can be computed. However, since the
solution to the general setup involves the same conceptual underpinnings, for
the sake of clarity we use a model of the neuron whose membrane potential
function is additively separable (i.e., $n=1$). The Spike Response Model (SRM),
introduced in \cite{gerstner2002}, is one such model. Although simple, the SRM
has been shown to be fairly versatile and accurate at modeling real biological
neurons \cite{jolivet2004}. The membrane potential, $P$, of the neuron, at the
present time is given by

\begin{equation}
\label{eq:srm}
 P = \sum\limits_{i \in \Gamma} w_{i}
          \sum\limits_{j \in \mathcal{F}_{i}} \xi_{i}(t^{I}_{i,j} - d_{i}) +
     \sum\limits_{k \in \mathcal F} \eta(t^{O}_{k})
\end {equation}

where $\Gamma$ is the set of synapses, $w_{i}$ is the weight of synapse $i$,
$\xi_{i}$ is the prototypical postsynaptic potential (PSP) elicited by a spike
at synapse $i$, $d_{i}$ is the axonal/synaptic delay, $t^{I}_{i,j}-d_{i}$ is
the time elapsed since the arrival of the $j^{th}$ most recent afferent
(incoming) spike at synapse $i$, and $\mathcal{F}_{i}$ is the potentially
infinite set of past spikes at synapse $i$. Likewise, $\eta$ is the
prototypical after-hyperpolarizing potential (AHP) elicited by an efferent
(outgoing) spike of the neuron, $t^{O}_{k}$ is the time elapsed since the
departure of the $k^{th}$ most recent efferent spike, and $\mathcal{F}$ is the
potentially infinite set of past efferent spikes of the neuron. The neuron
generated a spike whenever $P$ crosses the threshold $\Theta$ from below.

We make two additional assumptions: (i) the neuron has an absolute refractory
period that prohibits it from generating consecutive spikes closer than a given
bound $r$, and (ii) all input and output spikes that have aged past a given
bound $\Upsilon$ have no impact on the {\em present} membrane potential of the
neuron.

\begin{figure}[t!]
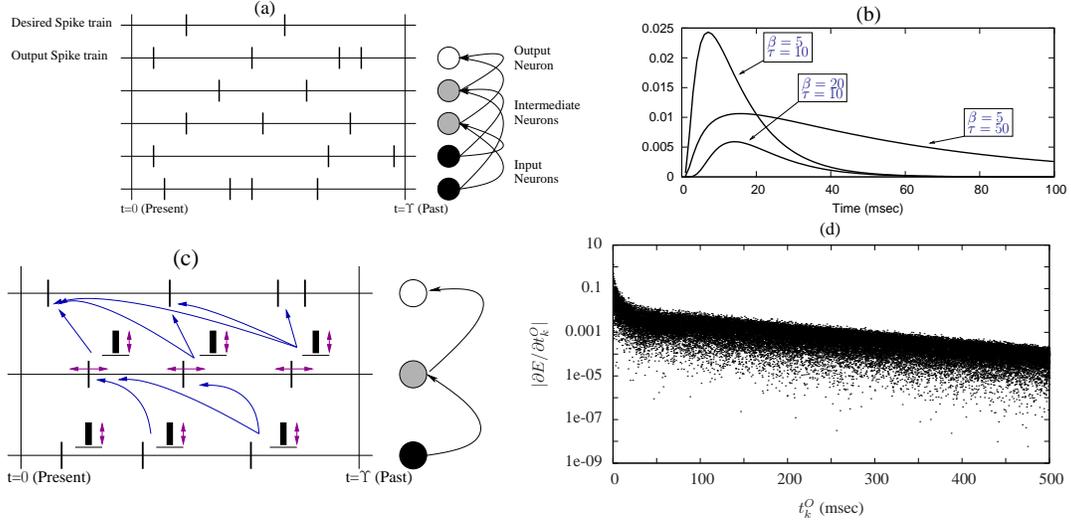

\centering
\resizebox{3.25in}{!}{\input framework.pstex_t}
\resizebox{2.25in}{!}{\input example_reefs_w_label.pstex_t}
\resizebox{2.5in}{!}{\input basic_idea.pstex_t}
\resizebox{3.0in}{!}{\input Output_delEdeltVSt.pstex_t}
\vspace{0.1in}
\caption{\setstretch{1}
(a) A feedforward network with two input neurons (shown in black), two
intermediate layer neurons (shown in gray) and one output neuron (shown in
white). The spike configuration in the bounded time window, $t=0$ (Present)
to $t=\Upsilon$ (in the past) is shown. Also shown is the desired output
spike train. Note that the desired and the output spike trains differ both in
their spike times as well as the number of spikes in the noted time window.
(b) The parameterized function $f_{\beta,\tau}(t)$ for various values of
$\beta$ and $\tau$.
(c) Assigning weights to spikes (denoted by the height of the bars) instead of
the corresponding synapse enables a perturbation analysis that derives the
effect of a change in the timing or the weight of a spike on the timing of
future spikes generated in the network. The weight perturbations are then
suitably accumulated at the synapse. The effect of perturbations (marked in
magenta) are computed for all input spike weights and intermediate spike times
and weights. They are not computed for input spike times since input spike
times are given and cannot be perturbed. The effects of the perturbations on
other spike times in the configuration space are marked in blue. Note that the
blue edges form a directed acyclic graph owing to causality.
(d) Simulation data demonstrating that gradient updates can have significant
values only immediately after the generation of a spike or the stipulation of a
desired spike at the output neuron. Scatter plot of the absolute value of
$\frac{\partial E}{\partial t^{O}_{k}}$ in log-scale plotted against
$t^{O}_{k}$. The values are drawn from 10,000 randomly generated pairs of
vectors ${\mathbf t^{O}}$ and ${\mathbf t^{D}}$.}
\label{fig:framework}
\end{figure}

The biological underpinnings of assumption (i) are well known. Assumption (ii)
is motivated by the following observations. It is generally accepted that all
PSPs and AHPs after an initial rise or fall, decay exponentially fast to the
resting potential. This, in conjunction with the existence of an absolute
refractory period, implies that for any given $\epsilon$ however small, there
exists an $\Upsilon$ such that the sum total effect of all spikes that have
aged past $\Upsilon$ can be bounded above by $\epsilon$ (see
\cite{banerjee2001}). Finally, observing that the biological neuron is a finite
precision device, we arrive at assumption (ii). The import of the assumptions
is that the size of $\mathcal{F}_{i}$ and $\mathcal{F}$ can now be bounded
above by $\lceil \Upsilon/r \rceil$. In essence, one has to merely look at a
bounded past to compute the present membrane potential of the neuron, and
moreover, there are only finitely many efferent and afferent spikes in this
bounded past. It helps to conceptualize the state of a network of neurons as
depicted in Figure~\ref{fig:framework}(a). The future spike trains generated by
the neurons in the network depend only on the future input spikes and the
spikes of all neurons in the bounded window $[0,\Upsilon]$.

We make two other changes. First, we shift the function $\xi_{i}$ to the right
so as to include the fixed axonal/synaptic delay. By so doing, we are relieved
of making repeated reference to the delay in the analysis. More precisely, what
was previously $\xi_{i}(t^{I}_{i,j} - d_{i})$ is now $\xi_{i}(t^{I}_{i,j})$,
with the new shifted $\xi_{i}$ satisfying $\xi_{i}(t)=0$ for $t<d_{i}$. The AHP
$\eta$ remains as before satisfying $\eta(t)=0$ for $t<0$. Second, and this has
major consequences, since our objective is to update the synaptic weights in an
online fashion, successive spikes on the same synapse can have potentially
different weights (assigned to the spike at its arrival at the synapse). We
account for this by assigning weights to spikes rather than synapses; we
replace $w_{i}$ by $w_{i,j}$. With these changes in place, we have

\begin{equation}
\label{eq:srm_new}
 P = \sum\limits_{i \in \Gamma} \sum\limits_{j \in \mathcal{F}_{i}}
      w_{i,j}\ \xi_{i}(t^{I}_{i,j}) +
     \sum\limits_{k \in \mathcal F} \eta(t^{O}_{k})
\end {equation}

\section{The Error Functional}

Having truncated the output spike train to a finite length vector of spike
times, we turn to the error functional. The problem, stated formally, is: given
two vectors of spike times, the output spike train $\langle t^{O}_{1},
t^{O}_{2}, \ldots, t^{O}_{N} \rangle$ and the desired spike train $\langle
t^{D}_{1}, t^{D}_{2}, \ldots, t^{D}_{M} \rangle$ of potentially differing
lengths, assign the pair a measure of disparity.

There have been several such measures proposed in the literature (see
\cite{victor1996,vanrossum2001,schreiber2003} for details). However, for
reasons that we delineate here, these measures do not fit our particular needs
well. First and foremost comes the issue of temporal asymmetry. As described
earlier, the effect of a spike on the potential of a neuron diminishes with age
in the long run, until it ceases altogether at $\Upsilon$. We prefer a measure
of disparity that focuses its attention more on the recent than the distant
past. If the output and desired spike trains align well in the recent past,
this is indicative of the synaptic weights being in the vicinity of their
respective desired values. A measure that does not suppress disparity in the
distant past will lead weight updates to overshoot. Second comes the issue of
the complex relationship between a spike train and its impact on the potential
of a neuron, which is the quantity of real interest. We prefer a measure that
makes this relationship explicit. Finally comes the issue of the ease with
which the measure can be manipulated. We prefer a measure that one can take the
gradient of, in closed form. We present a measure that possesses these
qualities.

We begin with a parameterized class of non-negative valued functions with shape
resembling PSPs.

\begin{equation}
f_{\beta,\tau}(t) = \frac{1}{\tau} e^{\frac{-\beta}{t}} e^{\frac{-t}{\tau}} \quad \textrm{ for} \quad\beta,\tau \ge 0\quad\textrm{ and }\quad t>\epsilon>0
\end{equation}

The functions are simplified versions of those in \cite{macgregor1977}.
Figure~\ref{fig:framework}(b) displays these functions for various values of
$\beta$ and $\tau$.

We set the putative impact of the vector of output spike times ${\mathbf t^{O}}
=\langle t^{O}_{1}, t^{O}_{2}, \ldots, t^{O}_{N} \rangle$ on a virtual
postsynaptic neuron to be $\sum_{i=1}^{N} f_{\beta,\tau}(t^{O}_{i})$, and
likewise for the vector of desired spike times ${\mathbf t^{D}} =\langle
t^{D}_{1}, t^{D}_{2}, \ldots, t^{D}_{M} \rangle$. Our goal is to assess the
quantity

\begin{equation}
\left( \sum_{i=1}^{M} f_{\beta,\tau}(t^{D}_{i}) - \sum_{i=1}^{N}
f_{\beta,\tau}(t^{O}_{i})\right)^2
\end{equation}

There are two paths we can pursue to
eliminate the dependence on the parameters $\beta,\tau$. The first is to set
them to particular values. However, reasoning that it is unlikely for a
presynaptic neuron to be aware of the shape of the PSPs of its postsynaptic
neurons, of which there may be several with differing values of $\beta,\tau$,
we follow the second path; we integrate over $\beta$ and $\tau$. Although
$\beta$ can be integrated over the range $[0,\infty)$, integrating $\tau$ over
the same range results in spikes at $\Upsilon$ having a fixed and finite
impact on the membrane potential of the neuron. To regain control over the
impact of a spike at $\Upsilon$, we integrate $\tau$ over the range
$[0,{\cal T}]$, for a reasonably large ${\cal T}$. By setting $\Upsilon$ to be
substantially larger that ${\cal T}$, we can make the impact of a spike at
$\Upsilon$ be arbitrarily small. We therefore have:

\begin{equation}
E({\mathbf t^{D}},{\mathbf t^{O}})=
\int_{0}^{\cal T} \int_{0}^{\infty}
\left( \sum_{i=1}^{M} f_{\beta,\tau}(t^{D}_{i}) - \sum_{i=1}^{N}
f_{\beta,\tau}(t^{O}_{i})\right)^2 d\beta d\tau
\end{equation}

Following a series of algebraic manipulations and noting that

\begin{equation}
\int_{0}^{\cal T} \int_{0}^{\infty}
\frac{1}{\tau}e^{\frac{-\beta}{t_1}}e^{\frac{-t_1}{\tau}} \times
\frac{1}{\tau}e^{\frac{-\beta}{t_2}}e^{\frac{-t_2}{\tau}} d\beta d\tau =
\frac{t_1 \times t_2}{(t_1+t_2)^2}e^{-\frac{t_1+t_2}{\cal T}}
\end{equation}

we get:

\begin{equation}
\label{eq:E}
E({\mathbf t^{D}},{\mathbf t^{O}})=
\sum_{i,j=1}^{M,M} \frac{t^{D}_{i} \times t^{D}_{j}}
                        {(t^{D}_{i} + t^{D}_{j})^2}
                   e^{-\frac{t^{D}_{i} + t^{D}_{j}}{\cal T}} +
\sum_{i,j=1}^{N,N} \frac{t^{O}_{i} \times t^{O}_{j}}
                        {(t^{O}_{i} + t^{O}_{j})^2}
                   e^{-\frac{t^{O}_{i} + t^{O}_{j}}{\cal T}} -
2\sum_{i,j=1}^{M,N} \frac{t^{D}_{i} \times t^{O}_{j}}
                         {(t^{D}_{i} + t^{O}_{j})^2}
                    e^{-\frac{t^{D}_{i} + t^{O}_{j}}{\cal T}}
\end{equation}

$E(\cdot)$ is a bounded from below and achieves its minimum value, 0, at
${\mathbf t^{O}}={\mathbf t^{D}}$. Computing the gradient of $E(\cdot)$ in
Eq~\ref{eq:E}, we get:

\begin{equation}
\label{eq:delEdelT}
\begin{split}
& \frac{\partial E}{\partial t^{O}_i}=\\
& 2\left(
         \sum_{j=1}^{N} \frac{t^{O}_{j}((t^{O}_{j} - t^{O}_{i})-
                      \frac{t^{O}_{i}}{{\cal T}}(t^{O}_{j} + t^{O}_{i}))}
                                   {(t^{O}_{j} + t^{O}_{i})^3}
                                   e^{-\frac{t^{O}_{j} + t^{O}_{i}}{{\cal T}}}
   -
         \sum_{j=1}^{M} \frac{t^{D}_{j}((t^{D}_{j} - t^{O}_{i})-
                      \frac{t^{O}_{i}}{{\cal T}}(t^{D}_{j} + t^{O}_{i}))}
                                   {(t^{D}_{j} + t^{O}_{i})^3}
                                   e^{-\frac{t^{D}_{j} + t^{O}_{i}}{{\cal T}}}
   \right)
\end{split}
\end{equation}

\section{Perturbation Analysis}

We now turn our attention to how perturbations in the weights and times of the
input spikes of a neuron translate to perturbations in the times of its output
spikes.\footnote{We remind the reader that weights are assigned to spikes and
not just to synapses to account for the online nature of synaptic weight
updates.} The following analysis applies to any neuron in the network, be it an
output or an intermediate layer neuron. However, we continue to refer to the
input and output spike times as $t^{I}_{i,j}$ and $t^{O}_{k}$ to keep the
nomenclature simple.

Consider the state of the neuron at the time of the generation of output spike
$t^{O}_{l}$. Based on the present spike configuration, we can write

\begin{equation}
\label{eq:base}
 \tilde{\Theta} =
   \sum\limits_{i \in \Gamma} \sum\limits_{j \in \mathcal{F}_{i}}
      w_{i,j}\ \xi_{i}(t^{I}_{i,j} - t^{O}_{l})
   + \sum\limits_{k \in \mathcal F} \eta(t^{O}_{k} - t^{O}_{l})
\end {equation}

Note that following definitions, $\xi_{i}$ returns the value 0 for all
$t^{I}_{i,j} < t^{O}_{l} + d_{i}$. Likewise $\eta$ returns the value 0 for all
$t^{O}_{k} < t^{O}_{l}$. In other words, we do not have to explicitly exclude
input/output spikes that were generated after $t^{O}_{l}$. Note also that we
have replaced the threshold $\Theta$ with $\tilde{\Theta}$. This reflects the
fact that we are missing the effects of all spikes that at the time of the
generation of $t^{O}_{l}$ had values less that $\Upsilon$ but are currently
aged beyond that bound. Since these are not quantities that we propose to
perturb, their effect on the potential can be considered a constant.

Had the various quantities in Eq~\ref{eq:base} been perturbed in the past, we
would have

\begin{equation}
\label{eq:perturb}
 \tilde{\Theta} =
   \sum\limits_{i \in \Gamma} \sum\limits_{j \in \mathcal{F}_{i}}
      (w_{i,j} + \Delta w_{i,j})\ \xi_{i}(t^{I}_{i,j} +\Delta t^{I}_{i,j}
                                        - t^{O}_{l} - \Delta t^{O}_{l})
   + \sum\limits_{k \in \mathcal F} \eta(t^{O}_{k} +\Delta t^{O}_{k}
                                       - t^{O}_{l} - \Delta t^{O}_{l})
\end {equation}

Combining Eq~\ref{eq:base} and Eq~\ref{eq:perturb} and using a first order
Taylor approximation, we get:

\begin{equation}
\label{eq:master}
\Delta t^{O}_{l}=
\frac{
   \sum\limits_{i \in \Gamma} \sum\limits_{j \in \mathcal{F}_{i}}
     \Delta w_{i,j} \xi_{i}(t^{I}_{i,j} - t^{O}_{l})
+
   \sum\limits_{i \in \Gamma} \sum\limits_{j \in \mathcal{F}_{i}}
     w_{i,j} \frac{\partial \xi_{i}}{\partial t}
                   \!\! \mid_{(t^{I}_{i,j} - t^{O}_{l})}
                   \Delta t^{I}_{i,j}
+
   \sum\limits_{k \in \mathcal F}
     \frac{\partial \eta}{\partial t}
                   \!\! \mid_{(t^{O}_{k} - t^{O}_{l})}
                   \Delta t^{O}_{k}
}
{
   \sum\limits_{i \in \Gamma} \sum\limits_{j \in \mathcal{F}_{i}}
     w_{i,j} \frac{\partial \xi_{i}}{\partial t}
                   \!\! \mid_{(t^{I}_{i,j} - t^{O}_{l})}
+
   \sum\limits_{k \in \mathcal F}
     \frac{\partial \eta}{\partial t}
                   \!\! \mid_{(t^{O}_{k} - t^{O}_{l})}
}
\end {equation}

We can now derive the final set of quantities of interest from
Eq~\ref{eq:master}:

\begin{equation}
\label{eq:delTdelW}
\frac{\partial t^{O}_{l}}{\partial w_{i,j}}=
\frac{
      \xi_{i}(t^{I}_{i,j} - t^{O}_{l})
+
   \sum\limits_{k \in \mathcal F}
     \frac{\partial \eta}{\partial t}
                   \!\! \mid_{(t^{O}_{k} - t^{O}_{l})}
                   \frac{\partial t^{O}_{k}}{\partial w_{i,j}}
}
{
   \sum\limits_{i \in \Gamma} \sum\limits_{j \in \mathcal{F}_{i}}
     w_{i,j} \frac{\partial \xi_{i}}{\partial t}
                   \!\! \mid_{(t^{I}_{i,j} - t^{O}_{l})}
+
   \sum\limits_{k \in \mathcal F}
     \frac{\partial \eta}{\partial t}
                   \!\! \mid_{(t^{O}_{k} - t^{O}_{l})}
}
\end {equation}

and

\begin{equation}
\label{eq:delTdelT}
\frac{\partial t^{O}_{l}}{\partial t^{I}_{i,j}}=
\frac{
     w_{i,j} \frac{\partial \xi_{i}}{\partial t}
                   \!\! \mid_{(t^{I}_{i,j} - t^{O}_{l})}
+
   \sum\limits_{k \in \mathcal F}
     \frac{\partial \eta}{\partial t}
                   \!\! \mid_{(t^{O}_{k} - t^{O}_{l})}
                   \frac{\partial t^{O}_{k}}{\partial t^{I}_{i,j}}
}
{
   \sum\limits_{i \in \Gamma} \sum\limits_{j \in \mathcal{F}_{i}}
     w_{i,j} \frac{\partial \xi_{i}}{\partial t}
                   \!\! \mid_{(t^{I}_{i,j} - t^{O}_{l})}
+
   \sum\limits_{k \in \mathcal F}
     \frac{\partial \eta}{\partial t}
                   \!\! \mid_{(t^{O}_{k} - t^{O}_{l})}
}
\end {equation}

The first term in the numerator of Eq~\ref{eq:delTdelW} and
Eq~\ref{eq:delTdelT} corresponds to the direct effect of a perturbation. The
second term corresponds to the indirect effect through perturbations in earlier
output spikes. The equations are a natural fit for an online framework since
the effects on earlier output spikes have previously been computed.

\section{Learning via Gradient Descent}

We now have all the ingredients necessary to propose a gradient descent based
learning mechanism. Stated informally, neurons in all layers update their
weights proportional to the negative of the gradient of the error functional.
In what follows, we specify the update for an output layer neuron and an
intermediate layer neuron that lies one level below the output layer. The
generalization to deeper intermediate layer neurons follows along similar
lines.

\subsection{Synaptic weight update for an output layer neuron}

In this case we would like to institute the gradient descent update $w_{i,j}
\longleftarrow w_{i,j} - \mu \frac{\partial E}{\partial w_{i,j}}$, where $\mu$
is the learning rate. However, since the $w_{i,j}$'s belong to input spikes in
the {\em past}, this would require us to reach back into the past to make the
necessary change. Instead, we institute a delayed update where the present
weight at synapse $i$ is updated to reflect the combined contributions from the
finitely many past input spikes in ${\mathcal {F}_{i}}$. Formally,

\begin{equation}
\label{eq:delw}
w_{i} \longleftarrow w_{i} -
\sum\limits_{j \in \mathcal{F}_{i}} \mu \frac{\partial E}{\partial w_{i,j}}
\end{equation}

The updated weight is assigned to the subsequent spike at the time of its
arrival at the synapse. $\frac{\partial E}{\partial w_{i,j}}$ is computed
using the chain rule (see Figure~\ref{fig:framework}(c)), with the constituent
parts drawn from Eq~\ref{eq:delEdelT} and Eq~\ref{eq:delTdelW} summed over the
finitely many output spikes in ${\mathcal F}$:

\begin{equation}
\label{eq:delEdelwij}
\frac{\partial E}{\partial w_{i,j}}=
   \sum\limits_{k \in \mathcal F}
\frac{\partial E}{\partial t^{O}_{k}}
\frac{\partial t^{O}_{k}}{\partial w_{i,j}}
\end{equation}

\subsection{Synaptic weight update for an intermediate layer neuron}

The update to a synaptic weight on an intermediate layer neuron follows along
identical lines to Eq~\ref{eq:delw} and Eq~\ref{eq:delEdelwij} with indices
$\langle i,j \rangle$ replaced by $\langle g,h \rangle$. The computation of
$\frac{\partial t^{O}_{k}}{\partial w_{g,h}}$, the partial derivative of the
$k^{th}$ output spike time of the {\em output} layer neuron with respect to the
weight on the $h^{th}$ input spike on synapse $g$ of the {\em intermediate}
layer neuron, is as follows. To keep the nomenclature simple, we assume that
the $j^{th}$ output spike of the intermediate layer neuron,
$t^{H}_{j}=t^{I}_{i,j}$ the $j^{th}$ input spike at the $i^{th}$ synapse of the
output layer neuron. Then, applying the chain rule (see
Figure~\ref{fig:framework}(c)) we have:

\begin{equation}
\label{eq:deltokdelwgh}
\frac{\partial t^{O}_{k}}{\partial w_{g,h}}=
   \sum\limits_{j \in \mathcal {F}_{i}}
\frac{\partial t^{O}_{k}}{\partial t^{I}_{i,j}}
\frac{\partial t^{H}_{j}}{\partial w_{g,h}}
\end{equation}

with the constituent parts drawn from Eq~\ref{eq:delTdelT} applied to the
output layer neuron and Eq~\ref{eq:delTdelW} applied to the intermediate layer
neuron, summed over the finitely many output spikes of the intermediate layer
neuron which are identically the input spikes in ${\mathcal {F}_{i}}$ of the
output layer neuron.

\subsection{A caveat concerning finite step size}

The earlier perturbation analysis is based on the assumption that infinitesimal
changes in the synaptic weights or the timing of the afferent spikes of a
neuron lead to infinitesimal changes in the timing of its efferent spikes.
However, since the gradient descent mechanism described above takes finite,
albeit small, steps, caution is warranted for situations where the step taken
is inconsistent with the underlying assumption of the infinitesimality of the
perturbations. There are two potential scenarios of concern. The first is when
a spike is generated somewhere in the network due to the membrane potential
just reaching threshold and then retreating. A finite perturbation in the
synaptic weight or the timing of an afferent spike can lead to the
disappearance of that efferent spike altogether. The perturbation analysis does
account for this by causing the denominators in Eq~\ref{eq:delTdelW} and
Eq~\ref{eq:delTdelT} to tend to zero (hence, causing the gradients to tend to
infinity). To avoid large updates, we set an additional parameter that capped
the length of the gradient update vector.
The second scenario is one where a finite perturbation leads to the
appearance of an efferent spike. Since there exists, in principle, an
infinitesimal perturbation that does not lead to such an appearance, the
perturbation analysis is unaware of this possibility. Overall, these scenarios
can cause $E(\cdot)$ to rise slightly at that timestep. However, since these
scenarios are only encountered infrequently, the net scheme decreases
$E(\cdot)$ in the long run.

\section{Experimental Validation}

The efficacy of the learning rule derived in the previous section hinges on two
factors: the ability of the spike timing based error to steer synaptic weights
in the ``correct'' direction, and the qualitative nature of the nonlinear
landscape of spike times as a function of synaptic weights, intrinsic to any
multilayer network. We evaluate these in order.

We begin with a brief description of the PSP and AHP functions that were used
in the simulation experiments. We chose the PSP $\xi$ and the AHP $\eta$ to
have the following forms (see \cite{macgregor1977} for details):

\begin {equation}
\xi(t) = \frac{1}{\alpha \sqrt{t}} \, e^{\frac{-\beta \alpha^{2}}{t}} \, e^{ \frac{-t}{\tau_{1}}} \times \mathcal{H}(t) \qquad \textrm { and }
\end{equation}
\begin{equation}
\eta(t) = -A \, e^{\frac{-t}{\tau_{2}}} \times \mathcal{H}(t)
\end{equation}

For the PSP function, $\alpha$ models the distance of the synapse from the
soma, $\beta$ determines the rate of rise of the PSP, and $\tau_{1}$ determines
how quickly it decays. $\alpha$ and $\beta$ are in dimensionless units. For the
AHP function, $A$ models the maximum drop in potential after a spike, and
$\tau_{2}$ controls the rate at which the AHP decays. $\mathcal{H}(t)$ denotes
the Heaviside step function: $\mathcal{H}(t) = 1$ for $t > 0$ and $0$
otherwise. All model parameters other than the synaptic weights were held fixed
through the experiments. In the vast majority of our experiments, we set
$\alpha=1.5$ for an excitatory synapse and $1.2$ for an inhibitory synapse,
$\beta=1$, $\tau_{1} = 20\,msec$ for an excitatory synapse and $10\,msec$ for
an inhibitory synapse. In all experiments, we set $A =1000$ and $\tau_{2} =
1.2\,msec$. A synaptic delay $d$ was randomly assigned to each synapse in the
range $[0.4, 0.9]\,msec$. The absolute refractory period $r$ was set to
$1\,msec$ and ${\cal T}$ was set to $150\,msec$. $\Upsilon$ was set to
$500\,msec$ which made the impact of a spike at $\Upsilon$ on the energy
functional negligible.

\subsection{Framework for testing and evaluation}

Validating the learning rule would ideally involve presentations of pairs of
input/desired output spike trains with the objective being that of learning the
transformation in an unspecified feedforward network of spiking neurons.
Unfortunately, as observed earlier, the state of our current knowledge
regarding what spike train to spike train transformations feedforward networks
of particular architectures and neuron models can implement, is decidedly
limited. To eliminate this confounding factor, we chose a witness based
evaluation framework. Specifically, we first generated a network, with synaptic
weights chosen randomly and then fixed, from the class of architecture that we
wished to investigate (henceforth called the witness network). We drove the
witness network with spike trains generated from a Poisson process and recorded
both the precise input spike train and the network's output spike train. We
then asked whether a network of the same architecture, initialized with random
synaptic weights, could learn this input/output spike train transformation
using the proposed synaptic weight update rule.

We chose a conservative criterion to evaluate the performance of the learning
process; we compared the evolving synaptic weights of the neurons of the
learning network to the synaptic weights of the corresponding neurons of the
witness network. Specifically, the disparity between the synaptic weights of a
neuron in the learning network and its corresponding neuron in the witness
network was quantified using the mean absolute percentage error (MAPE): the
absolute value of the difference between a synaptic weight and the ``correct''
weight specified by the witness network, normalized by the ``correct'' weight,
averaged over all synapses on that neuron. A MAPE of $1.0$ in the plots
corresponds to $100 \%$. Note that $100 \%$ is the maximum achievable MAPE when
the synaptic weights are lower than the ``correct'' weights.

There are several reasons why this criterion is conservative. Firstly, due to
the finiteness of the length of the recorded input/output spike train of the
witness network, it is conceivable that there exist other witness networks that
map the input to the corresponding output. If the learning network were to tend
toward one of these competing witness networks, one would erroneously deduce
failure in the learning process. Secondly, turning the problem of learning a
spike train to spike train transformation into one of learning the synaptic
weights of a network adds a degree of complexity; the quality of the learning
process now depends additionally on the characteristics of the input. It is
conceivable that learning is slow or fails altogether for one input spike train
while it succeeds for another. Notably, the two extreme classes of spike train
inputs, weak enough to leave the output neuron quiescent or strong enough to
cause the output neuron to spike at its maximal rate, are both noninformative.
In spite of these concerns, we found this the most objective and persuasive
criterion.

\subsection{Time of update}

The synaptic weight update rule presented in the previous section does not
specify a time of update. In fact, the synaptic weights of the neurons in the
network can be updated at any arbitrary sequence of time points. However, as
demonstrated here, the specific nature of one of the constituent parts of the
rule makes the update insignificantly small outside a particular window of
time.

Note that $\frac{\partial E}{\partial t^{O}_{k}}$, the partial derivative of
the error with respect to the timing of the $k$th efferent spike of the output
neuron, appears in the update formulas of all synapses, be they on the output
neuron or the intermediate neurons. We generated 10,000 random samples of pairs
of vectors ${\mathbf t^{O}}= \langle t^{O}_{1}, t^{O}_{2}, \ldots, t^{O}_{N}
\rangle$ and ${\mathbf t^{D}}= \langle t^{D}_{1}, t^{D}_{2}, \ldots, t^{D}_{M}
\rangle$ with $N$ and $M$ chosen independently and randomly from the range
$[1,10]$ and the individual spike times chosen randomly from the range
$[0,\Upsilon]$. As noted earlier, $\Upsilon$ and ${\cal T}$ were set to $500$
and $150\,msec$, respectively. We computed $\frac{\partial E}{\partial
t^{O}_{k}}$ for the individual spikes in each ${\mathbf t^{O}}$ according to
Eq~\ref{eq:delEdelT}. Figure~\ref{fig:framework}(d) presents a scatter plot
in log-scale of the absolute value of $\frac{\partial E}{\partial t^{O}_{k}}$
plotted against $t^{O}_{k}$, for the entire dataset. As is clear from the plot,
$|\frac{\partial E}{\partial t^{O}_{k}}|$ drops sharply with $t^{O}_{k}$.
Hence, the only time period during which the gradient update formulas can have
significant values is when at least one $t^{O}_{k}$ is small, that is,
immediately after the generation of a spike by the output neuron. The symmetric
nature of Eq~\ref{eq:delEdelT} would indicate that this is also true for the
timing of the desired spikes. We therefore chose to make synaptic updates to
the entire network soon after the generation of a spike by the output neuron or
the stipulation of a desired spike at the output neuron.

\subsection{Efficacy of the error functional -- single layer networks}

It is clear from Eq~\ref{eq:srm_new} that the spike train output of a neuron,
given spike train inputs at its various synapses, depends nonlinearly on its
synaptic weights. The efficacy of the proposed error functional hinges on how
reliably it can steer the synaptic weights of the learning network toward the
synaptic weights of the witness network, operating solely on spike time
disparities. This is best evaluated in a single layer network (i.e., a single
neuron with multiple synapses) since that eliminates the additional confounding
nonlinearities introduced by multiple layers.

Consider an update to the synapses of a learning neuron at any point in time.
Observe that since the update is based on the pattern of spikes in the finite
window $[0,\Upsilon]$, there are therefore uncountably many witness neurons
that could have generated that pattern. This class of witness neurons is even
larger if there are fewer desired spike times in $[0,\Upsilon]$. A gradient
descent update that steers the synaptic weights of the learning neuron in the
direction of any one of these potential witness neurons would constitute a
``correct'' update. It follows that when given a single witness neuron,
correctness can only be evaluated over the span of multiple updates to the
learning neuron.

\begin{figure}[t!]
\centering
\scalebox{0.71}{\includegraphics{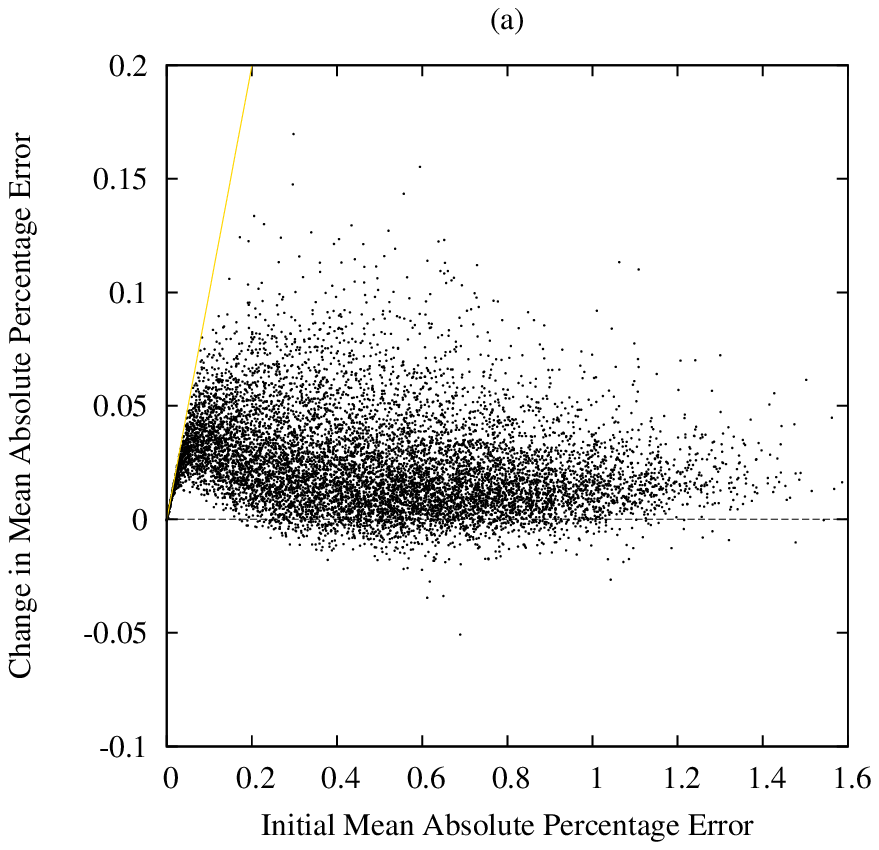}}
\scalebox{0.71}{\includegraphics{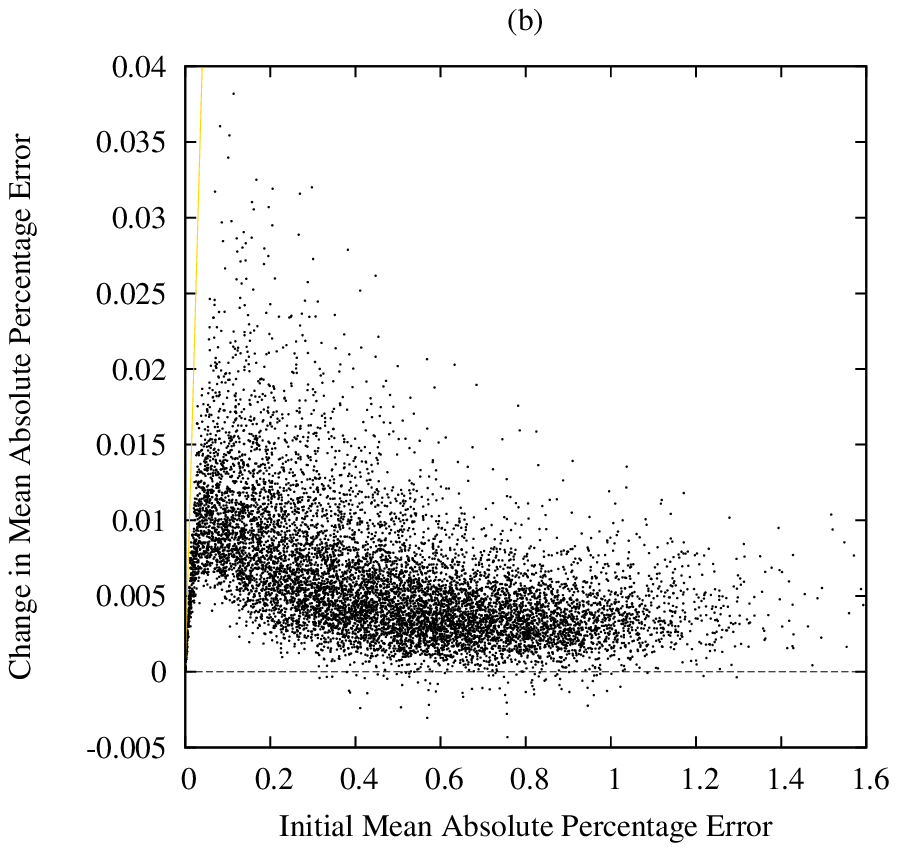}}
\scalebox{0.75}{\includegraphics{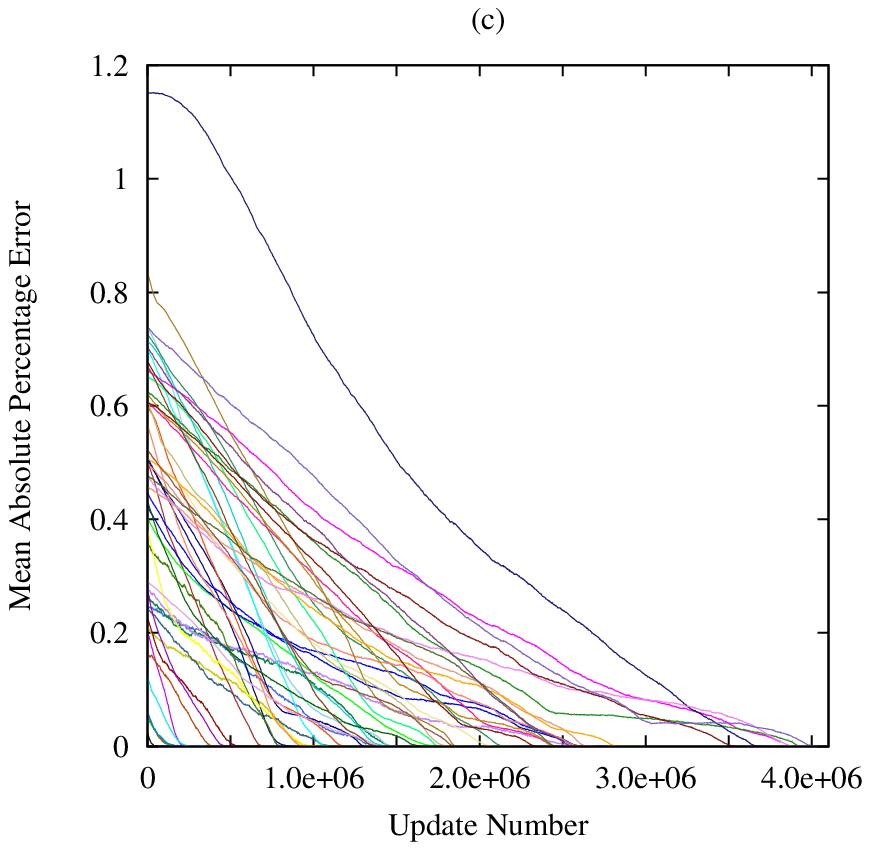}}
\scalebox{0.75}{\includegraphics{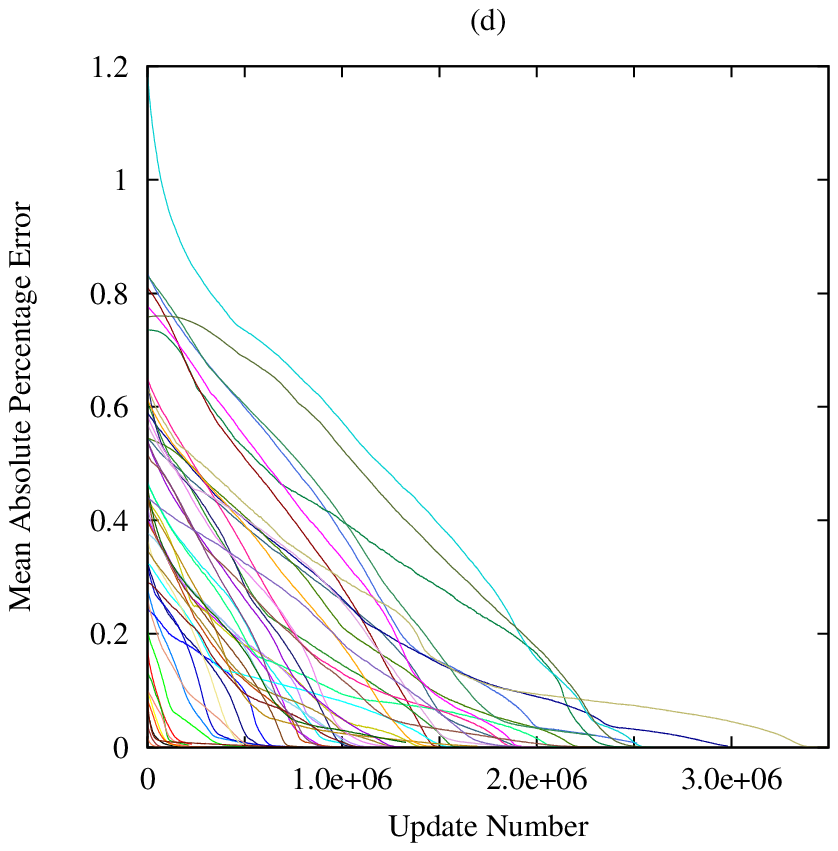}}
\caption{\setstretch{1}
Single neuron with $10$ synapses. (a) and (b) Scatter plot of initial
MAPE versus change in MAPE for $10,000$ witness-learning neuron pairs for a
bounded number of updates. The neurons in (a) were driven by homogeneous
Poisson spike trains and those in (b) by inhomogeneous Poisson spike trains.
Points on the yellow lines correspond to learning neurons that converged to
their corresponding witness neurons within the bounded number of updates. Note
that by definition, points cannot lie above the yellow lines.
(c) and (d) $50$ randomly generate witness-learning neuron pairs with learning
updates till convergence. Synapses on neurons in (c) are all excitatory and
those on neurons in (d) are 80\% excitatory and 20\% inhibitory. Each curve
corresponds to a single neuron. See text for more details regarding each
panel.}
\label{fig:expt-single-layer-10-syn}
\end{figure}

To obtain a global assessment of the efficacy landscape in its entirety, we
randomly generated $10,000$ witness-learning neuron pairs with $10$ excitatory
synapses each (the synaptic weights were chosen randomly from a range that made
the neurons spike between $5$ and $50$ Hz when driven by a $10$ Hz input) and
presented each pair with a randomly generated $10$ Hz Poisson input spike
train. Each learning neuron was then subjected to $50,000$ gradient descent
updates with the learning rate and cap set at small values. The initial versus
change in (that is, initial -- final) MAPE disparity between each learning and
its corresponding witness neuron is displayed as a scatter plot in
Figure~\ref{fig:expt-single-layer-10-syn}(a). Across the $10,000$ pairs, $9283$
($\approx 93\%$) showed improvement in their MAPE disparity. Furthermore, we
found a steady improvement of this percentage with increasing number of updates
(not shown here). Note that since the input rate was set to be the same across
all synapses, a rate based learning model would be expected to show improvement
in approximately $50\%$ of the cases.

A closer inspection of those learning neurons that did not show improvement
indicated the lack of diversity in the input spike patterns to be the cause.
We therefore ran a second set of experiments. Once again, as before, we
randomly generated $10,000$ witness-learning neuron pairs. Only this time,
input spike trains were drawn from an inhomogeneous Poisson process with the
rate set to modulate sinusoidally between $0$ and $10$ Hz at a frequency of $2$
Hz. The modulating rate was phase shifted uniformly for the $10$ synapses.
Surprisingly, after just $10,000$ gradient descent updates, $9921$ ($\approx
99\%$) neurons showed improvement, as displayed in
Figure~\ref{fig:expt-single-layer-10-syn}(b), indicating that with sufficiently
diverse input the error functional is {\em globally} convergent.

To verify the implications of the above finding with regard to the efficacy
landscape, we chose $50$ random witness-learning neuron pairs spread uniformly
over the range of initial MAPE disparities, and ran the gradient descent
updates until convergence (or divergence). Input spike trains were drawn from
the above described inhomogeneous Poisson process. All learning neurons
converged to their corresponding witness neurons as displayed in
Figure~\ref{fig:expt-single-layer-10-syn}(c).

The above experiments indicate that the error functional is globally convergent
to the ``correct'' weights when the synapses on the learning neuron are driven
by heterogeneous input. This finding can be related back to the nature of
$E(\cdot)$. As observed earlier, Eq~\ref{eq:E} makes $E(\cdot)$ nonnegative
with the global minima at ${\mathbf t^{O}}={\mathbf t^{D}}$. For synapses on
the learning and witness neuron pair to achieve this for all ${\mathbf t^{O}}$
and ${\mathbf t^{D}}$, they have to be identical. Furthermore, it follows from
Eq~\ref{eq:delEdelT} that a local minima, if one exists, must satisfy $N$
independent constraints for all ${\mathbf t^{O}}$ of length $N$. This is highly
unlikely for all ${\mathbf t^{O}}$ and ${\mathbf t^{D}}$ pairs generated by
distinct learning and witness neurons, particularly so if the input spike train
that drive these neurons is highly varied. Although, this does not exclude the
possibility of the sequence of updates resulting in a recurrent trajectory in
the synaptic weight space, the experiments indicate otherwise.

Finally, we conducted additional experiments with neurons that had a mix of
excitatory and inhibitory synapses with widely differing PSPs. In each of the
$50$ learning-witness neuron pairs, $8$ of the $10$ synapses were set to be
excitatory and the rest inhibitory. Furthermore, half of the excitatory
synapses were set to $\tau_1=80\,msec,\,\beta=5$, and half of the inhibitory
synapses were set to $\tau_1=100\,msec,\,\beta=50$ (modeling slower NMDA and
$\textrm{GABA}_B$ synapses, respectively). The results were consistent with the
findings of the previous experiments; all learning neurons converged to their
corresponding witness neurons as displayed in
Figure~\ref{fig:expt-single-layer-10-syn}(d).

\subsection{Nonlinear landscape of spike times as a function of synaptic
weights -- multilayer networks}

Having confirmed the efficacy of the learning rule for single layer networks,
we proceed to the case of multilayer networks. The question before us is
whether the spike time disparity based error at the output layer neuron,
appropriately propagated back to intermediate layer neurons using the chain
rule, has the capacity to steer the synaptic weights of the intermediate layer
neurons in the ``correct'' direction. Since the synaptic weights of any
intermediate layer neuron are updated based not on the spike time disparity
error computed at its output, but on the error at the output of the output
layer neuron, the overall efficacy of the learning rule depends on the
nonlinear relationship between synaptic weights of a neuron and output spike
times at a downstream neuron.

\begin{figure}[t!]
\centering
\scalebox{0.75}{\includegraphics{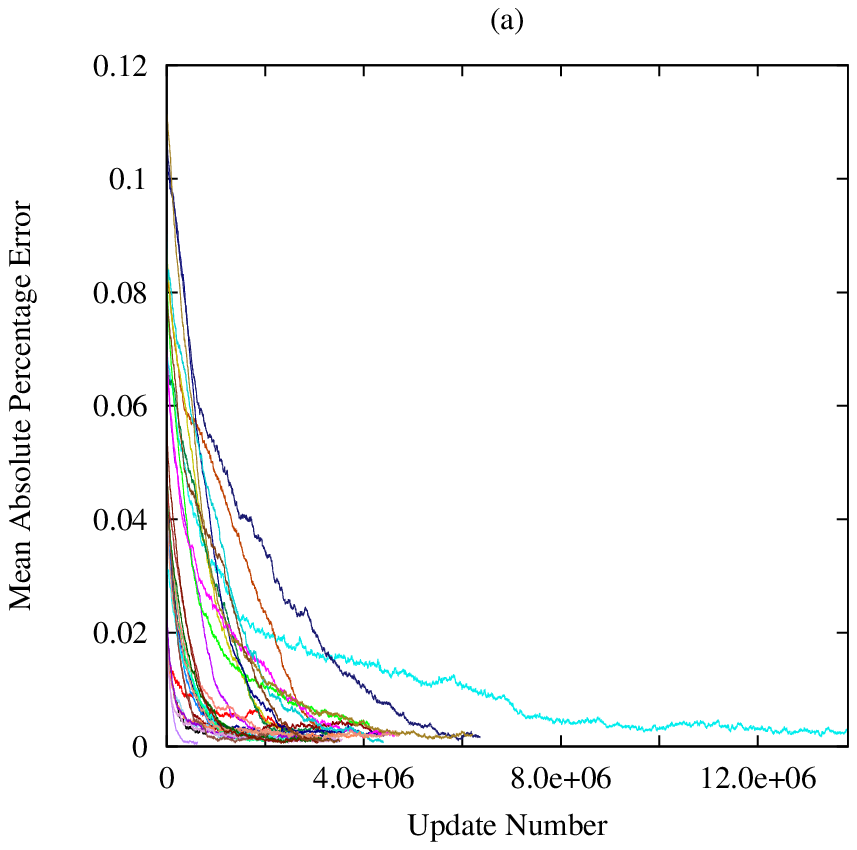}}
\scalebox{0.75}{\includegraphics{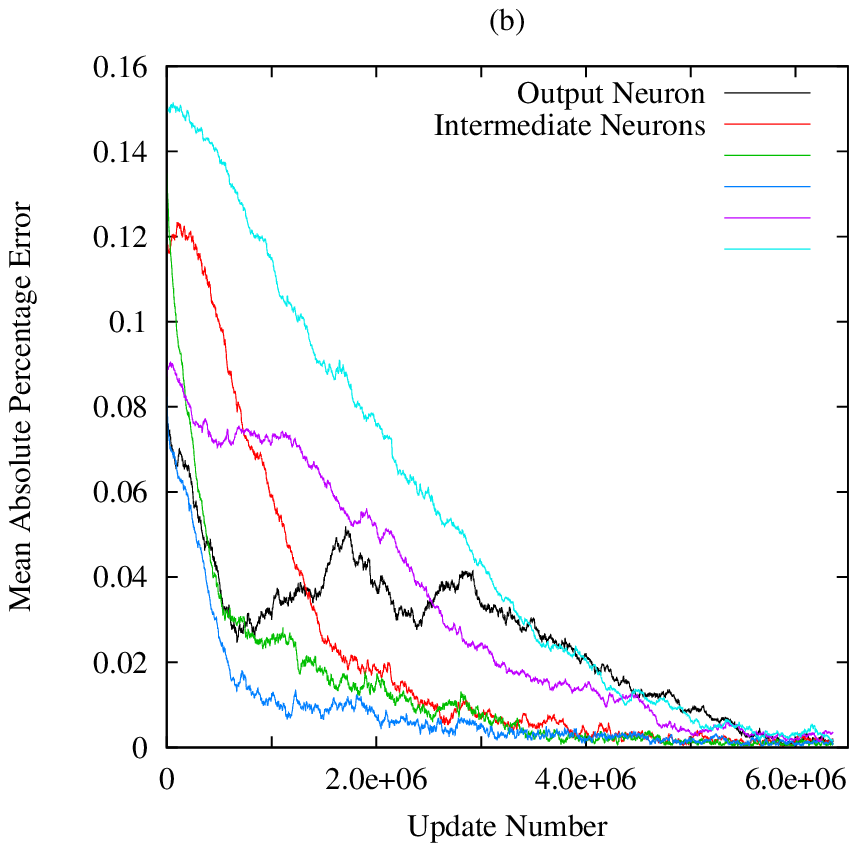}}
\scalebox{0.75}{\includegraphics{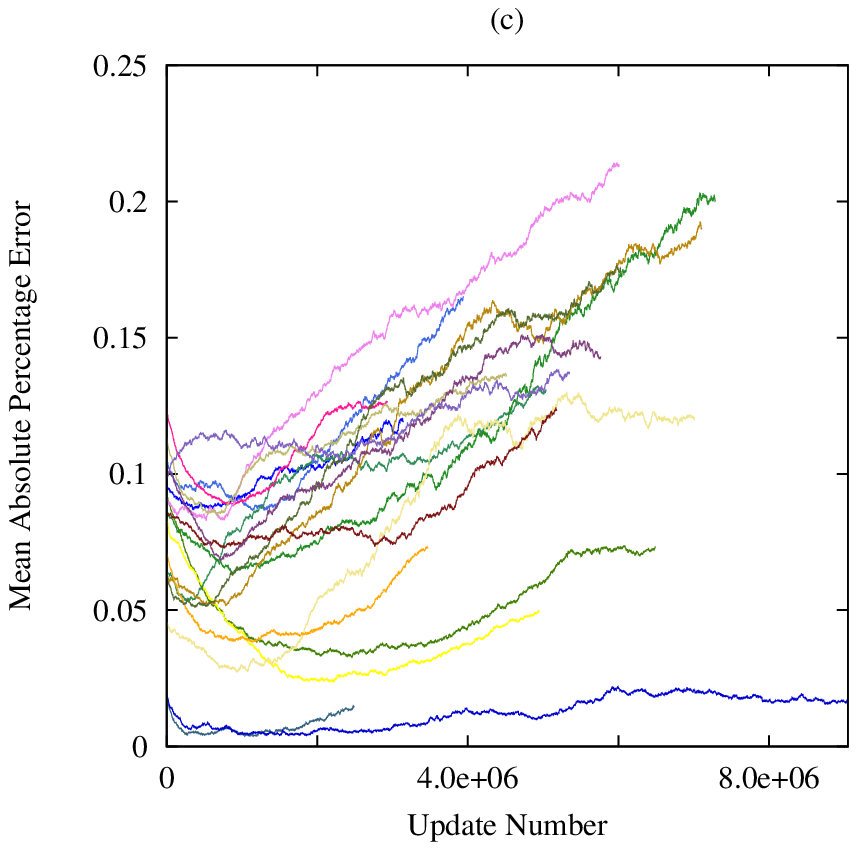}}
\scalebox{0.75}{\includegraphics{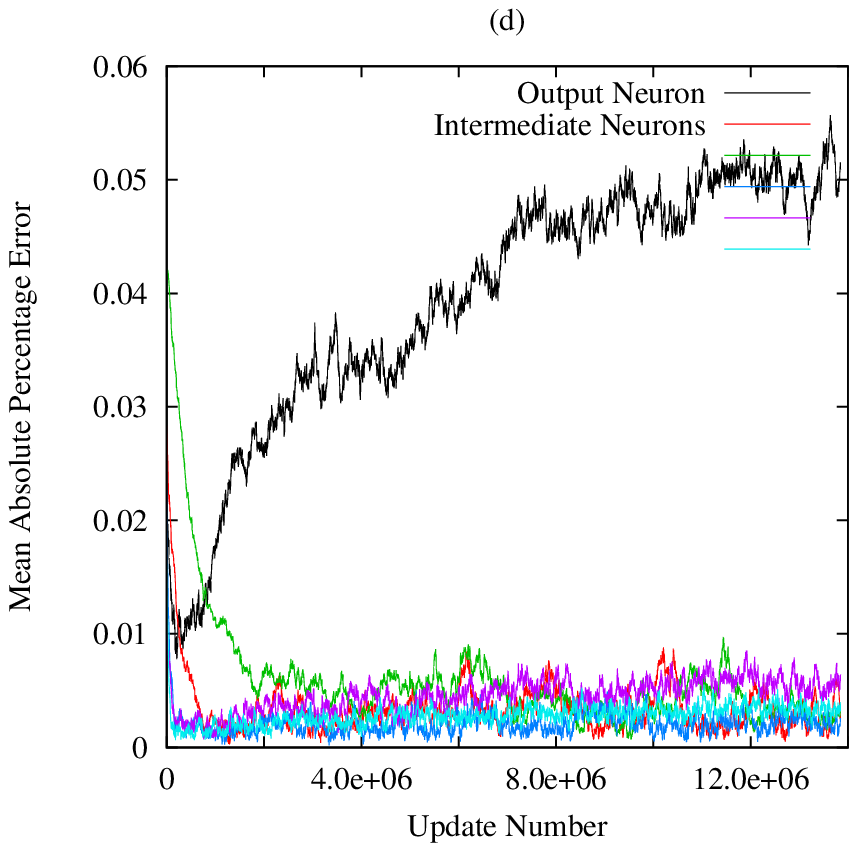}}
\caption{\setstretch{1}
Two layer networks with $30$ synapses ($5$ on each of $5$ intermediate
neurons and $5$ on the output neuron). (a) and (c) $50$ randomly generated
witness-learning network pairs with learning updates till convergence or
divergence. (a) $32$ of the networks converged and (c) the remaining $18$
networks diverged. Each curve corresponds to the average value of
the MAPE of the six neurons in the network. (b) and (d) Examples chosen from
(a) and (c) respectively showing the MAPE of all six neurons. See text for more
details regarding each panel.}
\label{fig:expt-two-layer-30-syn}
\end{figure}

We ran a large suite of experiments to assess this relationship. All
experiments were conducted on a two layer network architecture with five inputs
that drove each of five intermediate neurons which in turn drove an output
neuron. There were, accordingly, a sum total of $30$ synapses to train, $25$ on
the intermediate neurons and $5$ on the output neuron.

In the first set of experiments, as in earlier cases, we generated $50$ random
witness networks with all synapses set to excitatory. For each such witness
network we randomly initialized a learning network at various MAPE disparities
and trained it using the update rule. Input spike trains were drawn from an
inhomogeneous Poisson process with the rate set to modulate sinusoidally
between $0$ and $10$ Hz at a frequency of $2$ Hz, with the modulating rate
phase shifted uniformly for the $5$ inputs. The most significant insight
yielded by the experiments was that the domain of convergence for the weights
of the synapses, although fairly large, was not global as in the case of single
layer networks. This is not surprising and is akin to what is observed in
multilayer networks of sigmoidal neurons. Of the $50$ witness-learning network
pairs, $32$ learning networks converged to the correct synaptic weights, while
$18$ did not. Figure~\ref{fig:expt-two-layer-30-syn}(a) shows the average MAPE
disparity (averaged over the 5 intermediate and 1 output neuron) of the $32$
networks that converged to the ``correct'' synaptic weights.
Figure~\ref{fig:expt-two-layer-30-syn}(b) shows the MAPE of the six constituent
neurons of one of these $32$ networks; each curve in
Figure~\ref{fig:expt-two-layer-30-syn}(a) corresponds to six such curves.

\begin{figure}[t!]
\centering
\scalebox{0.75}{\includegraphics{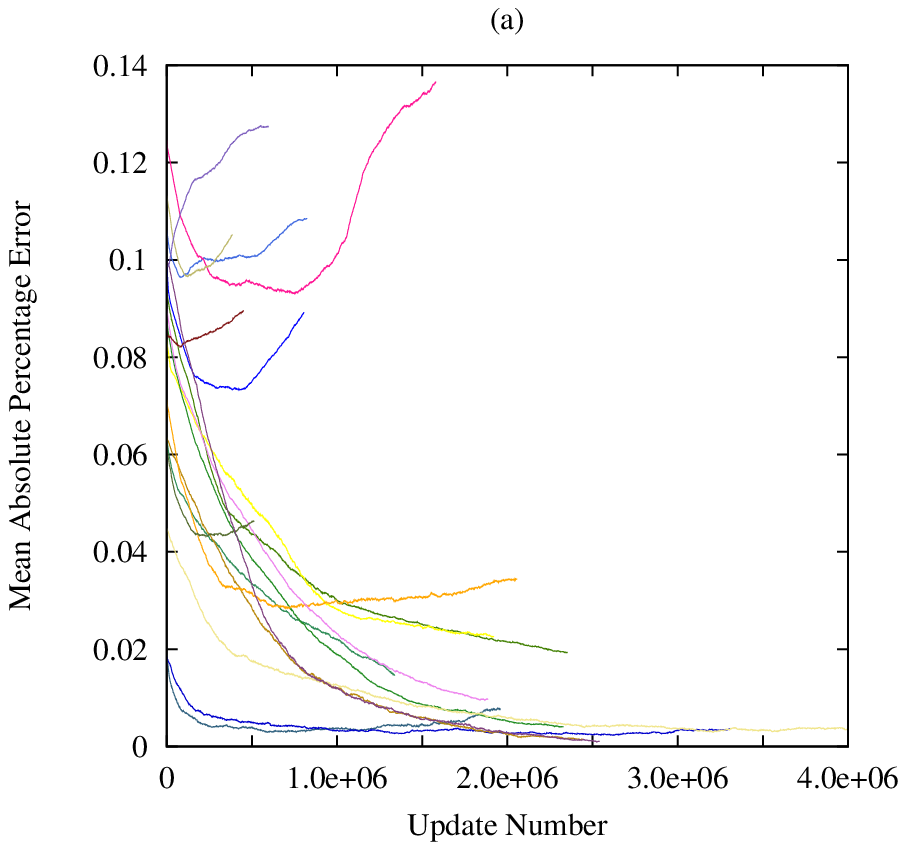}}
\scalebox{0.75}{\includegraphics{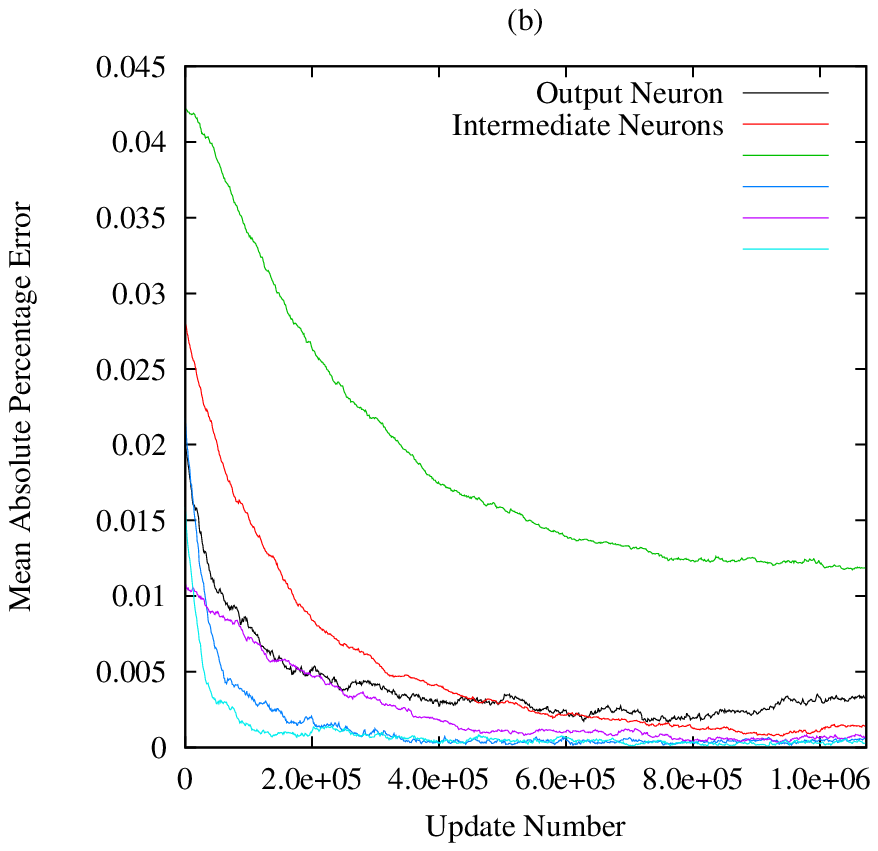}}
\caption{\setstretch{1}
(a) Witness-learning network pairs identical to those in
Figure~\ref{fig:expt-two-layer-30-syn}(c) driven by new, lower rate, input
spike trains. The maximum rate in the inhomogeneous Poisson process was reduced
from $10$ to $2$ Hz. The color codes for the specific learning networks are
left unchanged to aid visual comparison.
(b) The example network in Figure~\ref{fig:expt-two-layer-30-syn}(d) driven
by the new input spike train. Color codes are once again the same.}
\label{fig:expt-two-layer-30-syn-lr}
\end{figure}

Figure~\ref{fig:expt-two-layer-30-syn}(c) shows the average MAPE disparity
(averaged over the 5 intermediate and 1 output neuron) of the $18$ networks
that diverged. A closer inspection of the $18$ networks that failed to converge
to the correct synaptic weights indicated a myriad of reasons, not all implying
a definitive failure of the learning process. In many cases, all except a few
of the $30$ synapses converged. Figure~\ref{fig:expt-two-layer-30-syn}(d) shows
one such example where all synapses on intermediate neurons as well as three
synapses on the output neuron converged to their correct synaptic weights. For
synapses on networks that did not converge to the correct weights, the reason
was found to be excessively high or low pre/post synaptic spike rates, which as
was noted earlier are noninformative for learning purposes (incidentally, high
rates accounted for the majority of the failures in the experiments). To
elaborate, at high spike rates the tuple of synaptic weights that can generate
a given spike train is not unique. Gradient descent therefore cannot identify a
specific tuple of synaptic weights to converge to, and consequently the update
rule can cause the synaptic weights to drift in an apparently aimless manner,
shifting from one target tuple of synaptic weights to another at each update.
Not only do the synapses not converge, the error $E(\cdot)$ remains erratic and
high through the process. At low spike rates, gradients of $E(\cdot)$ with
respect to the synaptic weights drop to negligible values since the synapses in
question are not instrumental in the generation of most spikes at the output
neuron. Learning at these synapses can then become exceedingly slow.

To corroborate these observations, we ran a second set of experiments on the
$18$ witness-learning network pairs that did not converge. We reduced the
maximum modulating input spike rate from $10$ to $2$ Hz, i.e., input spike
trains were now drawn from an inhomogeneous Poisson process with the rate set
to modulate sinusoidally between $0$ and $2$ Hz at a frequency of $2$ Hz.
Figure~\ref{fig:expt-two-layer-30-syn-lr}(a) shows the average MAPE disparity
of the $18$ networks with the color codes for the specific networks left
identical to those in Figure~\ref{fig:expt-two-layer-30-syn}(c). Only $8$ of
the networks diverged this time. Figure~\ref{fig:expt-two-layer-30-syn-lr}(b)
shows the same network as in Figure~\ref{fig:expt-two-layer-30-syn}(d). This
time all synapses converged with the exception of one at an intermediate neuron
which displayed very slow convergence due to a low spike rate. We chose not to
further redress the cases that diverged in this set of experiments with new,
tailored, input spike trains to present a fair view of the learning landscape.

\begin{figure}[t!]
\centering
\scalebox{0.7}{\includegraphics{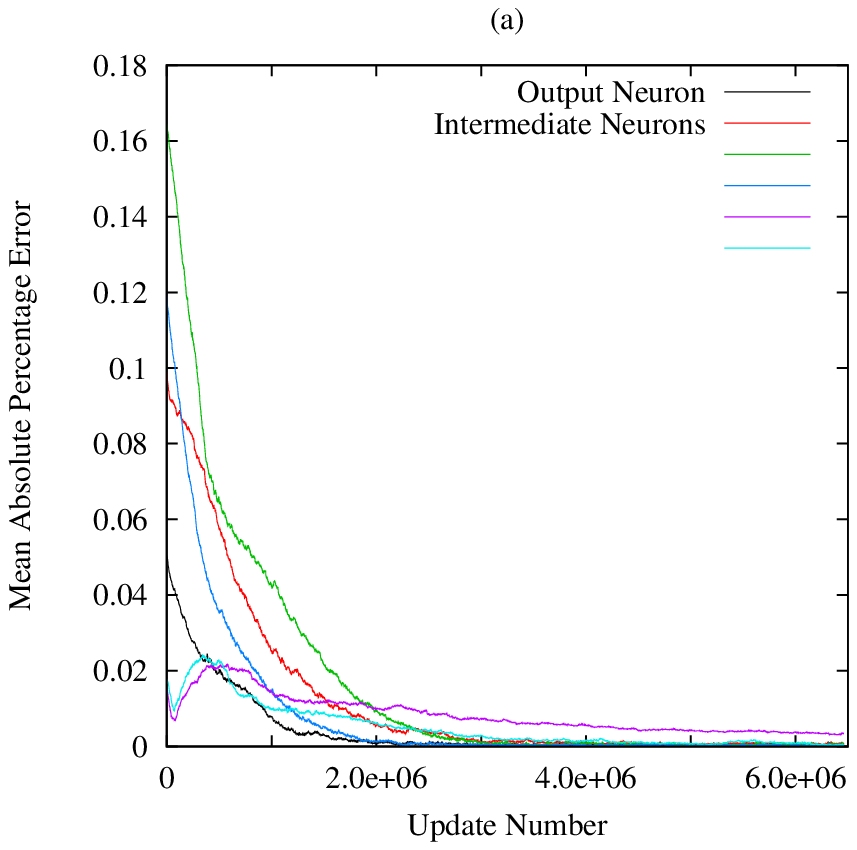}}
\scalebox{0.7}{\includegraphics{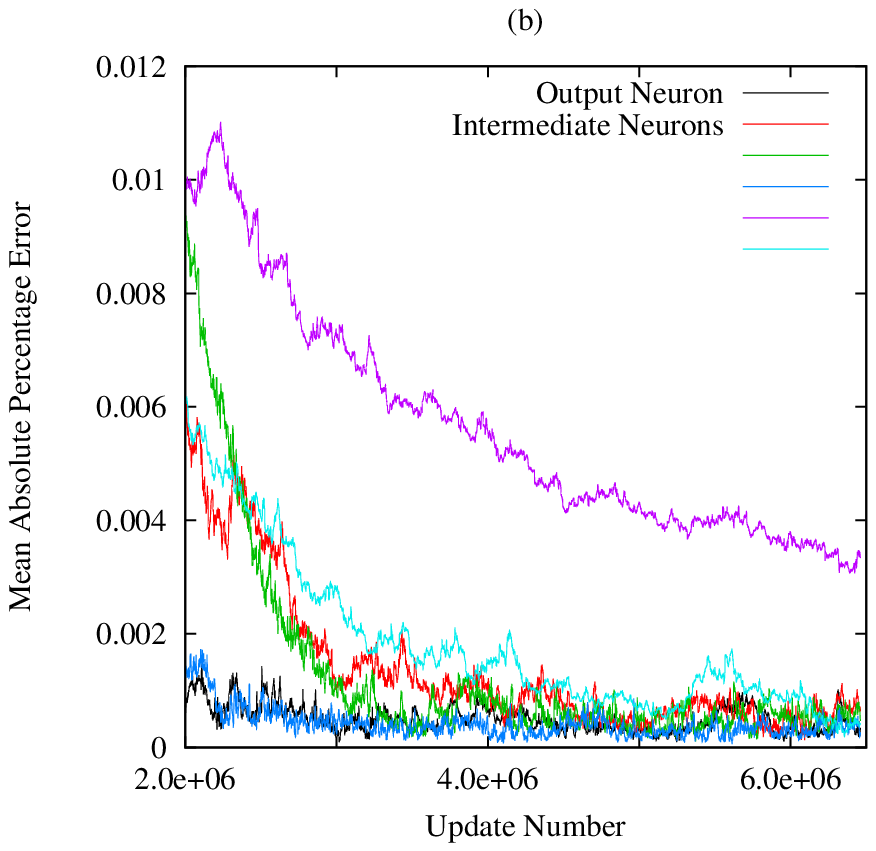}}
\caption{\setstretch{1}
(a) Two layer with $30$ synapses ($5$ on each of $5$ intermediate
neurons and $5$ on the output neuron) with two of the inputs and two of the
intermediate neurons set to inhibitory. Each curve corresponds to a single
neuron.
(b) Zoomed view of (a) showing slow convergence. See text for more
details regarding each panel.}
\label{fig:expt-two-layer-30-syn-exin}
\end{figure}

In our final set of experiments, we explored a network with a mix of excitatory
and inhibitory synapses. Specifically, two of the five inputs were set to
inhibitory and two of the five intermediate neurons were set to inhibitory.
The results of the experiments exhibited a recurring feature: the synapses on
the inhibitory intermediate neurons, be they excitatory or inhibitory,
converged substantially slower than the other synapses in the network.
Figure~\ref{fig:expt-two-layer-30-syn-exin}(a) displays an example of a network
that converged to the ``correct'' weights. Note, in particular, that the two
inhibitory intermediate neurons were initialized at a lower MAPE disparity as
compared to the other intermediate neurons, and that their convergence was
slow. The slow convergence is clearer in the the close-up in
Figure~\ref{fig:expt-two-layer-30-syn-exin}(b). The formal reason behind this
asymmetric behavior has to do with the range of values $\frac{\partial
t^{O}_{k}}{\partial t^{H}_{j}}$ takes for an inhibitory intermediate neuron as
opposed to an excitatory intermediate neuron, and its consequent impact on
Eq~\ref{eq:deltokdelwgh}. Observe that $\frac{\partial t^{O}_{k}}{\partial
t^{H}_{j}}$, following the appropriately modified Eq~\ref{eq:delTdelT}, depends
on the gradient of the PSP elicited by spike $t^{H}_{j}$ at the instant of the
generation of spike $t^{O}_{k}$ at the output neuron. The larger the gradient,
the greater is the value of $\frac{\partial t^{O}_{k}}{\partial t^{H}_{j}}$.
Typical excitatory (inhibitory) PSPs have a short and steep rising (falling)
phase followed by a prolonged and gradual falling (rising) phase. Since spikes
are generated on the rising phase of inhibitory PSPs, the magnitude of
$\frac{\partial t^{O}_{k}}{\partial t^{H}_{j}}$ for an inhibitory intermediate
neuron is smaller than that of an excitatory intermediate neuron. A remedy to
speed up convergence would be to compensate by scaling inhibitory PSPs to be
large and excitatory PSPs to be small, which, incidentally, is consistent with
what is found in nature.

\section{Discussion}

A synaptic weight update mechanism that learns precise spike train to spike
train transformations is not only of importance to testing forward models in
theoretical neurobiology, it can also one day play a crucial role in the
construction of brain machine interfaces. In this article, we have presented
such a mechanism formulated with a singular focus on the timing of spikes. The
rule is composed of two constituent parts, (a) a differentiable error
functional that computes the spike time disparity between the output spike
train of a network and the desired spike train, and (b) a suite of perturbation
rules that directs the network to make incremental changes to the synaptic
weights aimed at reducing this disparity. We have already explored (a), that
is, $\frac{\partial E}{\partial t^{O}_{k}}$ as defined in Eq~\ref{eq:delEdelT},
and presented its characteristic nature in Figure~\ref{fig:framework}(d). As
regards (b), when the learning network is driven by an input spike train that
causes all neurons, intermediate as well as output, to spike at moderate rates,
$\frac{\partial t^{O}_{l}}{\partial w_{i,j}}$ as defined in
Eq~\ref{eq:delTdelW} and $\frac{\partial t^{O}_{l}} {\partial t^{I}_{i,j}}$ as
defined in Eq~\ref{eq:delTdelT} can be simplified. Observe that when a neuron
spikes at a moderate rate, the past output spike times have a negligible AHP
induced impact on the timing of the current spike. Formally stated,
$\frac{\partial \eta}{\partial t}$ in Eq~\ref{eq:delTdelW} and
\ref{eq:delTdelT} are negligibly small for any output spike train with well
spaced spikes. Therefore,

\begin{equation}
\label{eq:_delTdelW}
\frac{\partial t^{O}_{l}}{\partial w_{i,j}} \approx
\frac{
      \xi_{i}(t^{I}_{i,j} - t^{O}_{l})
}
{
   \sum\limits_{i \in \Gamma} \sum\limits_{j \in \mathcal{F}_{i}}
     w_{i,j} \frac{\partial \xi_{i}}{\partial t}
                   \!\! \mid_{(t^{I}_{i,j} - t^{O}_{l})}
}
\end {equation}

and

\begin{equation}
\label{eq:_delTdelT}
\frac{\partial t^{O}_{l}}{\partial t^{I}_{i,j}} \approx
\frac{
     w_{i,j} \frac{\partial \xi_{i}}{\partial t}
                   \!\! \mid_{(t^{I}_{i,j} - t^{O}_{l})}
}
{
   \sum\limits_{i \in \Gamma} \sum\limits_{j \in \mathcal{F}_{i}}
     w_{i,j} \frac{\partial \xi_{i}}{\partial t}
                   \!\! \mid_{(t^{I}_{i,j} - t^{O}_{l})}
}
\end {equation}

The denominators in the equations above, as in Eq~\ref{eq:delTdelW} and
\ref{eq:delTdelT}, are normalizing constants that are strictly positive since
they correspond to the rate of rise of the membrane potential at the threshold
crossing corresponding to spike $t^{O}_{l}$. The numerators relate an
interesting story. Although both are causal, the numerator in
Eq~\ref{eq:_delTdelT} changes sign across the extrema of the PSP. Accumulated
in a chain rule, these make the relationship between the pattern of input and
output spikes and the resultant synaptic weight update rather complex.

Our experimental results have demonstrated that feedforward neuronal networks
can learn precise spike train to spike train transformations guided by the
weakest of supervisory signals, namely, the desired spike train at merely the
output neuron. Supervisory signals can of course be stronger, with the desired
spike trains of a larger subset of neurons in the network being provided. The
learning rule seamlessly generalizes to this scenario with the revised error
functional $E(\cdot)$ set as the sum of the errors with respect to each of the
supervising spike trains. What is far more intriguing is that the learning rule
generalizes to {\em recurrent} networks as well. This follows from the
observation that whereas neurons in a recurrent network cannot be partially
ordered, the spikes of the recurrent network in the bounded window
$[0,\Upsilon]$ can be partially ordered according to their causal structure
(see Figure~\ref{fig:framework}(c)), which then permits the application of the
chain rule. Learning in this scenario, however, seems to be at odds with the
sensitive dependence on initial conditions of the dynamics of a large class of
recurrent networks \cite{banerjee2006}, and therefore, the issue calls for
careful analysis.

\end{document}